\documentclass[11pt]{article}

\usepackage[final]{acl}
\usepackage{amsmath,amssymb}
\usepackage{algorithm}
\usepackage{algpseudocode}
\usepackage[most]{tcolorbox}
\usepackage{fvextra} 
\usepackage{booktabs}
\usepackage{adjustbox}
\usepackage{graphicx}
\usepackage{tcolorbox}
\usepackage{bbm}
\usepackage{dsfont}
\usepackage{times}
\usepackage{latexsym}

\usepackage[T1]{fontenc}

\usepackage[utf8]{inputenc}

\usepackage{microtype}

\usepackage{inconsolata}

\usepackage{graphicx}

\title{The Confident Liar: Diagnosing Multi-Agent Debate with Log-Probabilities and LLM-as-Judge}

\author{Ali Keramati, Justin Cheok , Jacob Horne and Mark Warschauer \\
University of California, Irvine\\
\texttt{\{a.kera,jcheok,jhorne1,markw\}@uci.edu} 
}

\begin{document}
\maketitle
\begin{abstract}
Multi-agent debate systems are typically evaluated only on whether the
final answer is correct, overlooking the quality of the intermediate
reasoning that debate is designed to produce. This paper studies the
relationship between three signals in multi-agent debate: token-level
log-probability distributions over reasoning tokens, LLM-as-judge rubric
scores assigned to those tokens, and final task accuracy. We examine
whether internal confidence signals predict externally evaluated reasoning
quality, and whether either signal aligns with task correctness, across
three domains: rubric-based scoring, mathematical reasoning, and factual
question answering. Our framework pairs a two-agent debate architecture
---a Constructor and an Auditor---with an
LLM-as-judge that scores each agent's reasoning along instruction
following, justification quality, and evidence grounding, together with a
critical-failure flag. Experiments in the rubric-scoring domain reveal a
consistent four-phase confidence trajectory and a substantial role
asymmetry: confidence aligns with judged reasoning quality roughly twice
as strongly for the Constructor as for the Auditor, and confidence-based detection of
critical reasoning failures is markedly more reliable for the Constructor
(AUROC 0.804) than for the Auditor (0.634). These findings motivate the
broader cross-domain investigation proposed in this paper.
\end{abstract}

\section{Introduction}
\label{sec:intro}
The rapid advancement of large language models (LLMs) has led to the 
emergence of multi-agent systems, where multiple specialized agents 
collaborate to solve complex tasks \cite{wu2023autogenenablingnextgenllm}. 
Such systems have been shown to improve performance, robustness, and 
consistency across a wide range of applications, including reasoning, 
planning, and automated decision-making 
\cite{parmar2025plangenmultiagentframeworkgenerating}. By decomposing 
tasks into role-specific subtasks, multi-agent frameworks enable more 
structured exploration of solution spaces compared to single-agent 
approaches \cite{10874775, han2026llmmultiagentsystemschallenges}. Among 
multi-agent interaction protocols, \emph{debate} has emerged as a 
particularly compelling mechanism: by eliciting both supporting and 
opposing arguments, debate encourages exploration of multiple reasoning 
pathways and can surface failure modes that remain hidden in a single 
trajectory \cite{du2023improvingfactualityreasoninglanguage}.

Despite their promise, multi-agent debate systems raise a fundamental 
evaluation challenge. In most settings, the only signal used to assess 
agent behavior is whether the final answer matches a reference 
answer, yet this binary signal fails to capture the quality, coherence, 
or reliability of the intermediate reasoning produced during debate. An 
agent may reach a correct answer through flawed reasoning, or produce a 
thoughtful argument that leads to a marginally incorrect conclusion. 
Evaluating only the endpoint discards the rich intermediate trace that 
debate was specifically designed to elicit. This is not merely a 
theoretical concern: as multi-agent pipelines grow in complexity, 
emergent failure modes may be invisible to accuracy-based metrics yet 
detectable in the structure of agents' reasoning \cite{wynn2025talkisntcheapunderstanding}.

A growing body of work addresses this need through \emph{LLM-as-judge} 
evaluation, in which a strong language model is prompted to score other 
model outputs along specified criteria \cite{zheng2023judging}. LLM-as-judge 
methods have become a scalable alternative to human evaluation for 
open-ended tasks and have been extended with rubric-based and 
fine-grained scoring protocols to assess intermediate reasoning rather 
than only final outputs \cite{ye2024flask, chan2024chateval}. Applied to 
debate, rubric-driven judging can evaluate whether an agent's argument 
is logically grounded, considers counterevidence, and engages 
substantively with the task, dimensions that final accuracy alone cannot 
capture \cite{chen2025multiagentasjudgealigningllmagentbasedautomated}. 
However, an important open question remains: \textit{do these external 
evaluations of reasoning quality reflect anything systematic about the 
model's own internal generation process?}

To answer this, we turn to \emph{confidence estimation} via token-level 
log-probabilities. Intuitively, if an agent follows a more coherent and 
evidentially grounded reasoning path, the model should assign higher 
probability mass to the tokens it generates along that path, yielding 
more concentrated logprob trajectories. Conversely, uncertain or 
contradictory reasoning may manifest as high entropy or spiky 
probability distributions over reasoning tokens 
\cite{kuhn2023semantic, kang2025scalable}. This framing opens a direct 
empirical question: \textit{to what extent do token-level log-probabilities 
of an agent's reasoning correlate with the quality of that reasoning as 
assessed by an external LLM judge, and does either signal align with 
downstream task accuracy?}

This paper proposes a systematic investigation of the 
relationship between these three signals---log-probability distributions 
over reasoning tokens, LLM-as-judge rubric scores applied to those 
tokens, and final task accuracy---across a diverse set of multi-agent 
debate tasks. Rather than focusing on any single domain, we study this 
triad of signals in a general multi-agent debate setting, using 
application domains such as rubric-based scoring, mathematical reasoning, 
and factual question answering as test beds. Our goal is to characterize 
how and when internal confidence signals align with externally assessed 
reasoning quality, which tasks and debate configurations drive the 
greatest divergence, and whether this divergence can be used 
diagnostically to improve multi-agent system design. This paper addresses the following key research questions:

\paragraph{RQ 1: Logprob Dynamics in Debate}
\begin{itemize}
    \item \textbf{RQ 1.1} How do token-level log-probability distributions 
    evolve across debate turns, and do agents expressing more 
    confident reasoning (higher logprobs) produce higher-quality 
    arguments as assessed by an LLM judge?
    \item \textbf{RQ 1.2} Can log-probability-based features predict 
    final task accuracy independently of the intermediate reasoning 
    content?
\end{itemize}

\paragraph{RQ 2: LLM-as-Judge Evaluation of Debate Reasoning}
\begin{itemize}
    \item \textbf{RQ 2.1}  How should rubric criteria be designed to 
    evaluate intermediate reasoning tokens in multi-agent debate, 
    and how consistent are LLM judges across different models and 
    protocols?
    \item \textbf{RQ 2.2} To what extent do LLM-as-judge scores on 
    intermediate reasoning correlate with final answer correctness 
    across diverse tasks?
\end{itemize}

\paragraph{RQ 3: Cross-signal Correlation and Diagnostics}
\begin{itemize}
    \item \textbf{RQ 3.1} Is there a systematic correlation between 
    logprob distributions, LLM-as-judge reasoning scores, and task 
    accuracy, and does this correlation vary across task types, 
    model families, or debate configurations?
    \item \textbf{RQ 3.2} Can divergence between internal confidence 
    signals and external reasoning quality assessments be exploited 
    to diagnose failure modes in multi-agent debate systems?
\end{itemize}

\section{Related Work}

\subsection{Multi-Agent Debate and Reasoning}

Multi-agent debate has been proposed as a mechanism to improve 
factuality, consistency, and robustness in LLM systems by having 
multiple models argue for and against candidate answers 
\cite{du2023improvingfactualityreasoninglanguage}. Empirical studies 
show that structured disagreement can reduce hallucination and improve 
reasoning on benchmarks spanning mathematics, logic, and 
question-answering 
\cite{chen2025multiagentasjudgealigningllmagentbasedautomated, 
wynn2025talkisntcheapunderstanding}. Beyond pairwise debate, 
multi-agent frameworks such as AutoGen support richer interaction 
topologies, enabling role specialization and more elaborate 
deliberation \cite{wu2023autogenenablingnextgenllm}. However, this 
added complexity creates new evaluation challenges: while debate 
generates rich intermediate reasoning traces, most prior work still 
evaluates these systems purely on final answer accuracy, leaving the 
quality of the intermediate argumentation unassessed 
\cite{han2026llmmultiagentsystemschallenges}. Our work directly 
targets this gap by pairing intermediate reasoning traces with 
both LLM-as-judge rubric scores and model-internal log-probability 
signals.

\subsection{LLM-as-Judge Evaluation}

Recent work has established LLM-as-judge as a practical paradigm for 
evaluating open-ended generation when reference answers are weak or 
unavailable, showing that strong proprietary models can often correlate 
well with human judgments on instruction-following and related tasks 
\cite{chiang2023can, li2023alpacaeval, zheng2023judging, fu2023gptscore, 
liu2023geval}. Beyond coarse pairwise or scalar judgments, a second 
line of work argues that evaluation should be more structured and 
interpretable: FLASK introduces fine-grained, skill-based assessment 
and shows that rubric-driven evaluation can improve both interpretability 
and reliability over skill-agnostic scoring \cite{ye2024flask}. More 
recent protocols incorporate chain-of-thought, multi-aspect scoring, 
and multi-agent evaluators such as PRD and ChatEval, all of which aim 
to elicit more reliable judgments \cite{li2023prd, chan2024chateval, 
jeong2024prepair}.

At the same time, a growing meta-evaluation literature has documented 
serious limitations: LLM judges exhibit verbosity and positional bias, 
limited self-consistency, and sensitivity to prompting and protocol 
design \cite{wang2023large, zeng2024llmbar}. REIFE demonstrates that 
protocol gains depend strongly on the base evaluator and dataset, 
and that reliable meta-evaluation requires diverse models and 
human-annotated testbeds \cite{liu2025reife}. Similarly, fine-tuned 
open-source judges such as JudgeLM, PandaLM, Auto-J, and Prometheus 
perform well in-domain yet fall short of frontier models in 
generalization and aspect-specific evaluation, suggesting they behave 
more like task-specific classifiers than general evaluators 
\cite{huang2025empirical}. Our work intersects with this literature 
by applying LLM-as-judge rubric evaluation specifically to 
intermediate reasoning tokens in debate, a setting where 
neither final accuracy nor coarse pairwise judgments adequately capture 
reasoning quality, and by investigating whether judge scores on 
reasoning correlate with the model's own internal confidence signals.

\subsection{Confidence Estimation and Uncertainty Quantification}

Confidence estimation in LLMs has emerged as an important complement 
to output evaluation, with prior work examining whether internal 
generation signals can indicate when model reasoning is trustworthy. 
Studies in calibration and uncertainty quantification show that 
neural probabilities are informative but not inherently well calibrated, 
so high model confidence does not always imply correctness 
\cite{desai2020calibration, kadavath2022language, kuhn2023semantic}. 
Even so, token-level probabilities remain one of the most direct 
intrinsic signals available during decoding, and a growing body of 
work uses log-probability- and entropy-based features to detect 
hallucination, factual errors, and uncertain generations 
\cite{liu2022token, manakul2023selfcheckgpt, mallen2023whennottrust}. 
In parallel, work on self-evaluation suggests that LLMs can sometimes 
report useful confidence judgments, but that verbalized confidence may 
differ from the model's underlying uncertainty, particularly for 
multi-step reasoning tasks \cite{kadavath2022language, 
mavi2025selfevaluatingllmsmultisteptasks}. Recent overviews argue for 
scalable uncertainty estimation methods that leverage intrinsic 
decoding-time signals alongside downstream evaluation criteria 
\cite{kang2025scalable}.

While this body of work has focused primarily on single-model generation 
and final-answer uncertainty, our work extends these ideas to the 
multi-agent debate setting. We study how logprob distributions behave 
not only at the level of final outputs but across the full sequence 
of intermediate reasoning tokens produced during multi-turn debate, 
and we ask whether these distributions correlate with external quality 
signals provided by an LLM judge.

\section{Methodology}
\label{sec:meth}

\begin{figure}[t]
  \includegraphics[width=\columnwidth]{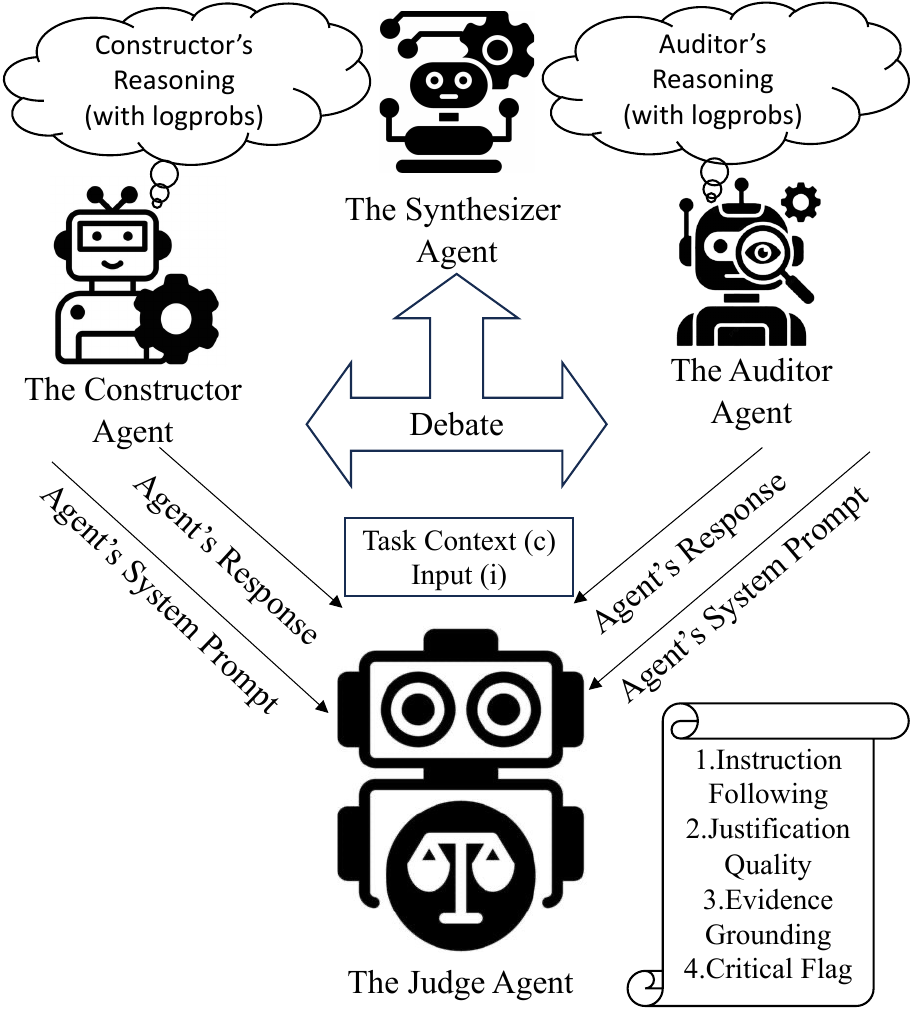}
  \caption{Overview of the proposed framework. A multi-agent debate 
  system generates structured reasoning over a task input; token-level 
  log-probabilities are extracted alongside each generation; a separate 
  LLM-as-judge module scores the reasoning; and all three signals are 
  correlated and analyzed.}
  \label{fig:methodology}
\end{figure}

Figure~\ref{fig:methodology} provides an overview of our framework, 
which operates across three stages: (1) a multi-agent debate system 
that generates structured reasoning over a task input, (2) a 
confidence extraction module that captures token-level log-probability 
trajectories from each agent's generation, and (3) an LLM-as-judge 
meta-evaluation module that scores each agent's intermediate reasoning 
against rubric-based criteria. Together, these stages produce three 
parallel signals for each debate instance (logprob features, judge 
scores, and downstream task accuracy) enabling us to study their 
joint distribution and mutual correlations. This design directly 
addresses \textbf{RQ 1}--\textbf{RQ 3} as outlined in 
Section~\ref{sec:intro}.

\subsection{Problem Setting}

Let $\mathcal{X}$ denote the set of task inputs and $\mathcal{C}$ the 
set of task contexts (e.g., rubric definitions, question prompts, 
grading criteria, or reference information), with $y^* \in \mathcal{Y}$ 
denoting the ground-truth label for each input. 
We study a general multi-agent debate setting in which two 
argumentative agents produce opposing or complementary arguments 
regarding a given input $x \in \mathcal{X}$ conditioned on task 
context $c \in \mathcal{C}$. A third agent, the 
Synthesizer, reads the completed debate transcript and 
produces the final task output $\hat{y}$.

Formally, for each task instance $(x, c, y^*)$, the debate system 
produces a transcript
\[
\tau(x, c) = (a,\, k),
\]
where $a$ denotes the first agent's argument and $k$ the second agent's 
rebuttal. Each argument is generated by a language model 
$\theta$ conditioned on the task input, context, and conversation 
history; the model simultaneously produces a sequence of token-level 
log-probabilities $L = (\ell_1, \dots, \ell_T)$ as an intrinsic 
confidence signal. Given a collection of debate instances 
$\{(a_i, k_i, L_i, q_i, y^*_i, \hat{y}_i)\}$, our objective is to 
analyze the pairwise and joint relationships among:
\begin{itemize}
    \item $L_i$: the log-probability trajectory of the generating agent,
    \item $q_i$: the LLM-as-judge score assigned to the agent's 
    reasoning, and
    \item $\mathbf{1}[\hat{y}_i = y^*_i]$: the final task accuracy 
    of the Synthesizer.
\end{itemize}

\subsection{Task Domains}
\label{sec:domains}

To ensure that our findings generalize beyond any single application, 
we instantiate the debate framework across three task domains that 
span different output structures and reasoning demands.

\paragraph{Rubric-Based Scoring.} The task is to assign a score to 
a textual input (e.g., an essay or short answer) according to an 
explicit rubric. The context $c$ consists of rubric trait definitions 
and scoring ranges. This domain is representative of settings where 
structured, multi-dimensional criteria govern evaluation, such as 
automated essay scoring \cite{Dikli_2006} or educational assessment. 
We use the ASAP dataset as a primary benchmark.\footnote{
\url{https://www.kaggle.com/c/asap-aes}}
For short-answer scoring, the ASAP-SAS dataset is also a relevant benchmark.\footnote{
\url{https://www.kaggle.com/competitions/asap-sas}}

\paragraph{Mathematical and Logical Reasoning.} The task involves 
solving a multi-step reasoning problem where the ground truth is a 
discrete answer. The context $c$ consists of the problem statement 
and any relevant constraints. This domain tests whether confidence 
signals are predictive when reasoning involves precise, verifiable 
intermediate steps, and whether the judge's assessment of argument 
quality aligns with mathematical correctness. We use GSM8K as a 
primary benchmark for this setting.\footnote{
\url{https://github.com/openai/grade-school-math}}

\paragraph{Factual Question Answering.} The task is to answer a 
factual question, where the ground truth is a specific entity or 
short phrase. The context $c$ may include retrieved passages or 
background knowledge. This domain tests the framework in a setting 
where agent arguments must appeal to factual evidence rather than 
structural criteria, providing a complementary signal to the rubric 
and reasoning domains. We use Natural Questions as a primary 
benchmark for this setting.\footnote{
\url{https://ai.google.com/research/NaturalQuestions/}}

\subsection{Agent Roles and Task-Specific Instantiation}
\label{sec:agents}

We define two abstract, domain-agnostic agent roles that 
capture the functional purpose of structured disagreement without 
presupposing the nature of the task:

\begin{itemize}
    \item \textbf{The Constructor} produces a primary response: 
    a candidate answer, solution, score, or position, together 
    with the reasoning that supports it. The Constructor is 
    designed to commit to a direction and develop it as fully 
    and coherently as possible, making its reasoning explicit 
    and traceable.
    \item \textbf{The Auditor} reads the Constructor's output 
    and produces a critical second response. Rather than simply 
    agreeing or disagreeing, the Auditor is tasked with 
    \emph{independently examining} the reasoning for errors, 
    gaps, unsupported claims, or overlooked alternatives, and 
    must provide its own evidence-based justification for any 
    challenge it raises.
\end{itemize}

These two roles preserve the core property of debate that motivates 
our study, that structured disagreement produces richer and more 
diverse reasoning traces than single-agent generation, while being 
sufficiently abstract to admit meaningful instantiation across all 
three task domains. A third agent, the \textbf{Synthesizer}, reads 
the completed exchange and produces the final task output $\hat{y}$. 
The Synthesizer is held constant across all domains and is excluded 
from the reasoning quality and confidence analyses, as its output 
is evaluated directly against the ground truth $y^*$. Table~\ref{tab:agent_roles} summarizes how the Constructor and 
Auditor are concretely instantiated in each task domain.

\begin{table*}[t]
\centering
\small
\begin{tabular}{p{2.2cm} p{3.8cm} p{4.2cm} p{4.2cm}}
\toprule
\textbf{Domain} & \textbf{Task Input $x$} & \textbf{Constructor} 
& \textbf{Auditor} \\
\midrule
Rubric-Based Scoring 
& Essay or short answer + rubric trait definition 
& \textit{Advocate}: argues for a specific score range by 
citing supporting evidence from the text; does not assign 
a final score 
& \textit{Critic}: challenges the Advocate's reading by 
identifying weaknesses, omissions, or rubric misalignments 
in the text \\
\addlinespace
Mathematical Reasoning 
& Multi-step problem statement + constraints 
& \textit{Solver}: proposes a complete solution path with 
explicit intermediate steps; commits to a final answer 
& \textit{Verifier}: independently checks each reasoning 
step for correctness, identifies the first error if present, 
and proposes an alternative derivation if the solution 
is flawed \\
\addlinespace
Factual Question Answering 
& Question + optional retrieved passages 
& \textit{Proponent}: argues for a specific candidate 
answer by citing evidence from the context or background 
knowledge; explains why competing answers are less 
supported 
& \textit{Challenger}: questions the Proponent's evidence, 
raises alternative candidate answers, and argues for the 
most strongly supported alternative \\
\bottomrule
\end{tabular}
\caption{Task-specific instantiation of the Constructor and 
Auditor roles across the three task domains. The Synthesizer 
agent is held constant across all domains and is not shown.}
\label{tab:agent_roles}
\end{table*}

\subsubsection{Role Design Principles}

Several design choices are shared across all instantiations and 
are motivated by the goals of the study.

\paragraph{No final answer from Constructor or Auditor.} In the 
rubric scoring and QA domains, neither the Constructor nor the 
Auditor is permitted to produce a final task output directly. 
This constraint ensures that their generations consist 
\emph{entirely} of reasoning tokens, making the log-probability 
trajectories we extract (Section~\ref{sec:logprobs}) reflective 
of argumentative reasoning rather than label decoding. In the 
math domain, the Constructor is an exception: it must produce a 
final numerical answer as part of its solution path, since the 
answer is inseparable from the derivation. However, the 
\emph{Verifier} is still prohibited from directly confirming or 
denying the answer without showing its own working.

\paragraph{Independence of the Auditor.} The Auditor is 
explicitly instructed not to simply restate the Constructor's 
reasoning with superficial modifications. In the math domain, 
the Verifier must rederive relevant steps independently before 
issuing a judgment. In the scoring and QA domains, the Auditor 
must cite specific textual or factual evidence distinct from 
that used by the Constructor. This independence constraint 
is essential for ensuring that the two agents' log-probability 
distributions reflect genuinely different reasoning paths, 
enabling meaningful comparison.

\paragraph{Structured output format.} All agents are prompted 
to organize their responses with explicit labeled sections 
(e.g., \textit{Claim}, \textit{Evidence}, \textit{Reasoning}, 
\textit{Conclusion}), adapted to each domain. This structure 
serves two purposes: it makes dimension-level LLM-as-judge 
evaluation (Section~\ref{sec:judge}) more reliable by providing 
clear anchors for scoring, and it allows us to align 
log-probability windows (Section~\ref{sec:logprobs}) with 
specific functional phases of the response in future work.

\paragraph{Adversarial vs. collaborative framing.} In the 
rubric and QA domains, the Constructor and Auditor are framed 
as \emph{adversarial}: the Auditor is incentivized to find 
flaws. In the math domain, the Verifier is framed as 
\emph{cooperative but skeptical}: its goal is to verify 
correctness rather than to find a flaw at all costs, since 
an incorrect verification is itself a failure mode we wish to 
detect. This distinction means that the expected relationship 
between Constructor and Auditor log-probability trajectories 
may differ across domains, which we treat as an empirical 
question addressed in Section~\ref{sec:correlation}.

\subsubsection{Prompt Design}

Full system prompts for all agent instantiations across the 
three domains are provided in \textbf{Appendix~\ref{app:prompts}}. 
All prompts share a common template structure consisting of: 
(i) a role preamble that defines the agent's persona and 
objective, (ii) behavioral constraints specifying what the 
agent must and must not do, (iii) the task input and context, 
(iv) the prior turn(s) of the debate when applicable, and 
(v) an output format specification. Role preambles are the 
only component that varies across agents; all other components 
are held as consistent as possible to isolate the effect of 
role assignment on generation behavior and log-probability 
distributions.

\subsection{Confidence Signals from Token Log-Probabilities}
\label{sec:logprobs}

We operationalize internal model confidence using token-level 
log-probabilities obtained during generation, addressing 
\textbf{RQ 1.1} and \textbf{RQ 1.2}. For a generated response of 
$T$ tokens, the model produces a log-probability at each decoding 
step:
\[
\ell_t = \log p(t_t \mid t_{<t},\, x, c),
\]
where $x$ is the task input, $c$ is the task context, and $t_{<t}$ 
denotes preceding tokens. The resulting sequence 
$L = (\ell_1, \dots, \ell_T)$ forms a log-probability trajectory 
over the full response.

\subsubsection{Window-Based Segmentation}

Rather than summarizing $L$ with a single global statistic, we 
extract contiguous sub-sequences to examine how confidence evolves 
across different phases of an agent's generation. This temporal 
decomposition allows us to test, for instance, whether opening 
claims or concluding statements are generated with systematically 
different confidence than the middle of the response. We use two 
complementary windowing strategies:

\paragraph{Fixed-length windows.} For window size $k$:
$$
W_{\text{first}}(k) = (\ell_1, \dots, \ell_k),
$$
$$
W_{\text{last}}(k)  = (\ell_{T-k+1}, \dots, \ell_T).
$$

\paragraph{Percentage-based windows.} To normalize across responses 
of varying length, we define windows as a fraction 
$\alpha \in (0,1]$ of the total response:

$$
W_{\text{first}}(\alpha) = (\ell_1, \dots, \ell_{\lfloor \alpha T 
\rfloor}),
$$
$$
W_{\text{last}}(\alpha)  = (\ell_{T - \lfloor \alpha T \rfloor 
+ 1}, \dots, \ell_T).
$$

\subsubsection{Statistical Aggregation}

For each window $W$, we compute the following summary statistics:

\paragraph{Mean and median.}
The mean reflects overall token likelihood, while the median provides
a robust central-tendency estimate that is less sensitive to outlier 
tokens:
\[
\mu_W = \frac{1}{|W|}\sum_{\ell \in W} \ell, 
\qquad 
\tilde{\mu}_W = \mathrm{median}(W).
\]

\paragraph{Minimum, maximum, and range.} $\min(W)$, $\max(W)$, and $\mathrm{range}_W = \max(W) - \min(W)$ bound 
the extent of confidence variation within the segment.

\paragraph{Variance and standard deviation.}
These quantify the volatility of the generation process within a window, 
which we hypothesize may indicate argumentative uncertainty:
\[
\sigma_W^2 = \frac{1}{|W|}\sum_{\ell \in W} (\ell - \mu_W)^2, 
\qquad 
\sigma_W = \sqrt{\sigma_W^2}.
\]

\paragraph{Trajectory slope.} 
We fit a linear regression to the log-probability sequence over the 
window:
\[
\ell_t \approx \beta_W\,t + b_W.
\]
The slope $\beta_W$ captures directional trends: $\beta_W > 0$ indicates 
growing model confidence across the segment (e.g., the model becomes 
more certain as the argument develops), while $\beta_W < 0$ indicates 
declining confidence.

\paragraph{Entropy-based aggregation.} 
Beyond summary statistics of scalar log-probabilities, we also compute 
the token-level entropy of the model's full output distribution at each 
decoding step:
\[
H_t = -\sum_{v \in V} p(v \mid t_{<t}, x, c)\, \log p(v \mid t_{<t}, x, c),
\]
and aggregate the sequence $\{H_t\}_{t=1}^{T}$ over each window $W$ 
using the same statistics defined above (mean, median, variance, 
standard deviation, range, and slope), yielding entropy analogues such 
as $\mu^H_W$ and $\sigma^H_W$. This provides a complementary view of 
uncertainty that captures the spread of the model's full predictive 
distribution rather than only the probability assigned to the chosen 
token
\cite{kuhn2023semantic}.

\subsection{LLM-as-Judge Meta-Evaluation of Reasoning}
\label{sec:judge}

Because the debating agents sometimes generate open-ended argumentative 
reasoning, their outputs cannot be evaluated with reference-based 
metrics. We therefore introduce a secondary evaluation stage in 
which a separate language model judges the quality of each agent's 
reasoning along rubric-based dimensions, addressing \textbf{RQ 2.1} 
and \textbf{RQ 2.2}.

\subsubsection{Prompt Reconstruction}

For each agent response, we reconstruct the complete prompt context 
the agent originally received, consisting of: (i) the agent's system 
instructions describing its role and behavioral constraints, 
(ii) the task context $c$, (iii) the task input $x$, and (iv) the 
agent's generated response. Supplying this full context enables the 
evaluator to assess both role adherence and the appropriateness 
of evidence use within the specific task.

\subsubsection{Evaluation Dimensions}

The meta-evaluator scores each response along three dimensions that 
are designed to be applicable across all three task domains 
(Section~\ref{sec:domains}):

\paragraph{Instruction Following.} Whether the agent maintained 
its assigned role throughout, avoided 
prohibited behaviors (e.g., declaring a final answer directly), 
and respected the task constraints specified in its system prompt.

\paragraph{Justification Quality.} Whether claims are supported 
by explicit reasoning steps that coherently link evidence to 
conclusions, and whether the agent engages substantively with 
the task rather than generating vague or generic statements.

\paragraph{Evidence Grounding.} Whether the argument references 
concrete, specific information from the task input, textual 
passages, numerical values, logical premises, or retrieved 
facts, rather than appealing to unsupported generalizations.

\subsubsection{Scoring Protocol}

Each dimension is scored on a three-point ordinal scale 
($1 = \text{Low},\ 2 = \text{Medium},\ 3 = \text{High}$), 
assigned independently. The evaluator also raises a 
\textbf{critical issue flag} when the response contains a 
severe failure that invalidates the reasoning, including 
hallucinated evidence, major internal contradictions, 
role-constraint violations, or incoherent output.

We compute an aggregate reasoning quality score:
\[
Q_1 = s_{\text{instruction}} + s_{\text{justification}} 
+ s_{\text{evidence}}.
\]
If a critical failure is detected, the aggregate score is 
overridden:
\[
Q =
\begin{cases}
0 & \text{if critical issue is present}, \\
Q_1 & \text{otherwise},
\end{cases}
\]
yielding a composite score $Q \in [0, 9]$.

\subsubsection{Judge Consistency Analysis}

To assess the reliability of the LLM judge itself we run each evaluation prompt through multiple state-of-the-art 
models (e.g., GPT-5.5, Claude Opus 4.7, Gemini 3 Pro) and 
compute inter-judge agreement using Krippendorff's $\alpha$ and 
rank correlation. We also run the same judge model multiple times 
under non-zero temperature to estimate intra-judge consistency. 
Dimension-level agreement is reported separately to identify which 
aspects of reasoning quality are most reliably assessable by 
automated judges, directly addressing \textbf{RQ 2.1}.

\subsection{Cross-Signal Correlation Analysis}
\label{sec:correlation}

The central empirical goal of this work is to characterize the 
joint distribution of logprob features, judge scores, and task 
accuracy. This section describes how we operationalize this 
analysis to address \textbf{RQ 3.1} and \textbf{RQ 3.2}.

\subsubsection{Pairwise Correlation}

For each pair of signals, we compute Pearson and Spearman correlation 
coefficients across all debate instances in a given domain. We use 
$\mu_W$ and $\sigma_W$ to denote the mean and standard deviation of an 
agent's log-probability trajectory aggregated over a window $W$, as 
defined in Section~\ref{sec:logprobs}; unless otherwise stated, $W$ is 
taken to be the full response (i.e., $W = L$). Specifically, we examine:

\begin{itemize}
  \item $\rho(Q,\, \mu_W)$: correlation between judge score and the mean 
  log-probability of the agent's response.
  \item $\rho(Q,\, \sigma_W)$: correlation between judge score and the 
  log-probability standard deviation (does more certain generation 
  correspond to higher-quality reasoning?).
  \item $\rho(\mathbb{1}[\hat{y} = y^*],\, Q)$: correlation between judge 
  score and task accuracy (does high-quality intermediate reasoning 
  predict correct final answers?).
  \item $\rho(\mathbb{1}[\hat{y} = y^*],\, \mu_W)$: direct 
  logprob--accuracy correlation, bypassing the judge entirely.
\end{itemize}

\subsubsection{Divergence Detection}

Beyond global correlations, we identify instances where the 
three signals \emph{diverge}, as these cases are diagnostically 
most informative (\textbf{RQ 3.2}). We define three divergence 
conditions:
\begin{itemize}
  \item \textbf{High-confidence, low-quality reasoning:} $\mu_W$ is high 
  but $Q$ is low. These cases suggest the model generates with high 
  internal certainty while producing reasoning the judge deems weak---a 
  potential hallucination or overconfidence signature.
  \item \textbf{High-quality reasoning, incorrect answer:} $Q$ is high 
  but $\hat{y} \neq y^*$. These cases suggest that good intermediate 
  argumentation does not always suffice for task success, pointing to 
  failures in the Synthesizer's integration of the debate.
  \item \textbf{Low-confidence, correct answer:} $\mu_W$ is low but 
  $\hat{y} = y^*$. These cases suggest the model can succeed despite 
  internally uncertain reasoning, motivating caution in using logprobs 
  alone as a quality proxy.
\end{itemize}

We analyze the prevalence of each divergence type across task 
domains and debate roles, and we qualitatively inspect 
high-divergence instances to characterize their failure modes.

\subsubsection{Stratification and Ablations}

To assess whether correlations are stable or task-specific, we 
stratify all analyses by: (i) task domain 
(Section~\ref{sec:domains}), (ii) debate role, (iii) model family and scale, and (iv) window position 
(first vs. last window of the response). We also ablate the 
aggregation statistic (mean, variance, slope) to determine which 
logprob features are most predictive of judge scores and accuracy, 
and we test whether trajectory slope adds predictive information 
beyond mean log-probability alone.

\section{Experiments and Results}

This section examines how token-level confidence behaves over the course 
of multi-agent debate and how that behavior relates to externally judged 
reasoning quality. We instantiate the framework in the rubric-based 
scoring domain using the ASAP dataset: Constructor and Auditor responses 
are generated by GPT-4o-mini with token-level log-probabilities recorded 
during decoding, and each response is independently scored by a 
GPT-5-mini judge along instruction following, justification quality, and 
evidence grounding, together with a binary critical-failure flag. The 
analysis proceeds in two stages.

\subsection{Confidence Trajectories Across Reasoning}

\begin{figure}[t]
  \centering
  \includegraphics[width=\columnwidth]{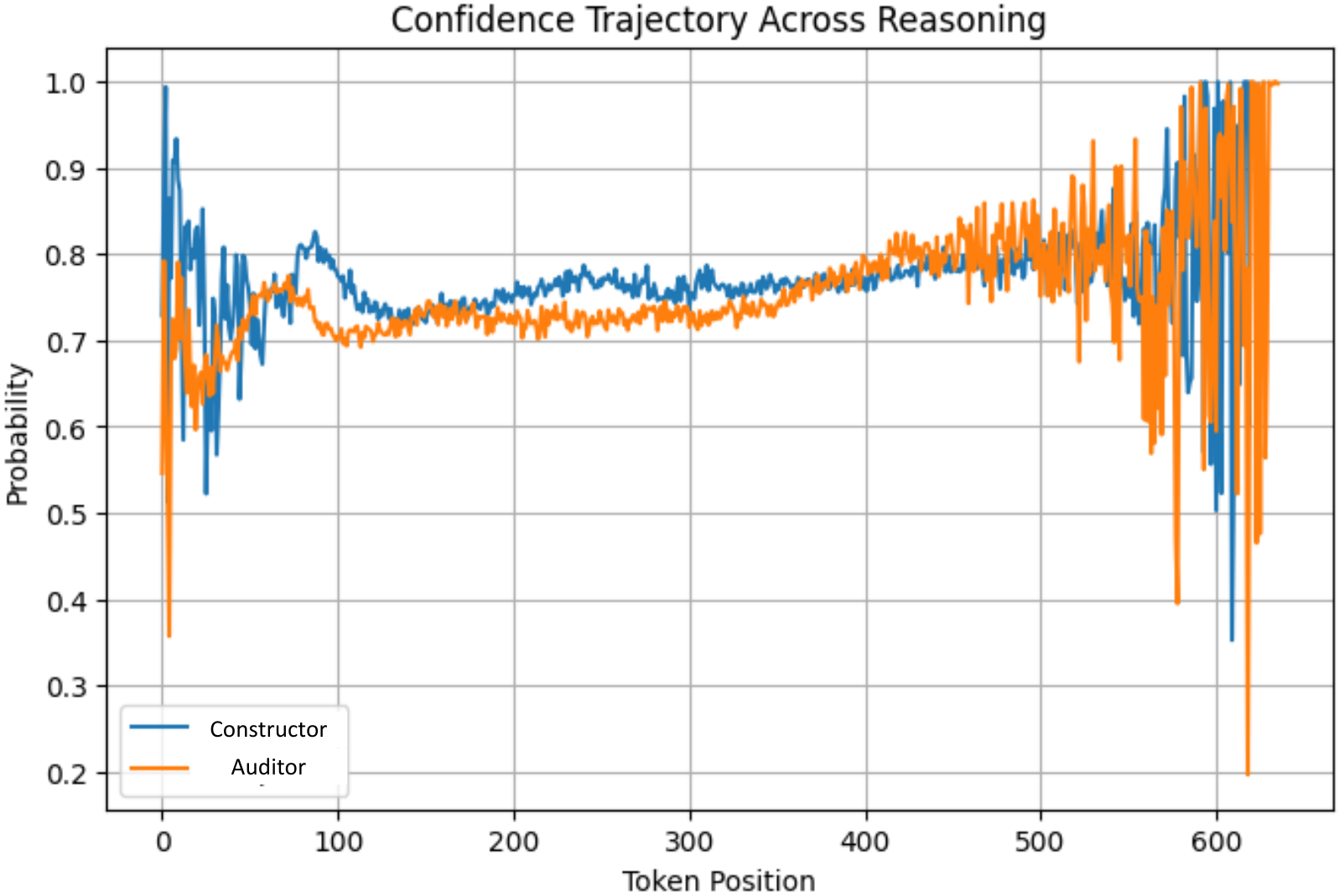}
  \caption{Mean token-level confidence trajectories for Constructor and 
  Auditor responses. Both agents follow a four-phase pattern---high-confidence 
  opening, sharp early decline, stable mid-response plateau, and volatile 
  late-response region.}
  \label{fig:trajectory}
\end{figure}

Figure~\ref{fig:trajectory} plots the mean token-level confidence 
trajectory of the Constructor and Auditor agents across reasoning. 
Responses open at high confidence, undergo a sharp decline within the 
first 50 tokens as substantive reasoning begins, settle into a stable 
plateau through the middle of the response, and become increasingly 
volatile near the end of generation. The replication of this four-phase 
pattern under a multi-agent debate setting suggests that the structural 
shape of confidence dynamics is not an artifact of any single task or role 
framing, but a more general property of how debating agents allocate 
certainty over a generation.

Two features of the trajectory are particularly informative. First, the 
role asymmetry between Constructor and Auditor is visible directly in the 
trajectory rather than only in aggregate statistics: across the plateau 
region (roughly tokens 100--400), the Constructor maintains consistently 
higher token probability than the Auditor, with a stable gap of 
approximately 0.05. This is consistent with the interpretation that 
supportive reasoning unfolds along a more committed and predictable path, 
while adversarial reasoning navigates a wider space of candidate critiques. 
Second, the late-response region (tokens 550+) exhibits substantially 
greater volatility than the rest of the trajectory, with both agents 
oscillating sharply between near-certain and low-confidence tokens. This 
tail behavior, which is partially obscured when trajectories are 
summarized by length-normalized averages, raises the possibility that the 
closing portion of a debate response carries diagnostic information that 
has so far been underutilized.

\subsection{Cross-Signal Correlations Across Reasoning Dimensions}

Beyond the macro-trajectory structure, we summarize the strength of the 
relationship between intrinsic confidence signals and externally judged 
reasoning quality, addressing RQ 1.1 and RQ 3.2. 
Table~\ref{tab:rq-summary} aggregates results across both ASAP essay sets, 
with each cell reporting the strongest correlation observed for the 
corresponding role--dimension pair after a sweep over the window-based 
confidence features defined in Section~\ref{sec:meth}.

Three patterns emerge from the aggregated results. First, every 
role--dimension pair exhibits a positive and nontrivial alignment between 
confidence and judged quality, supporting the central premise of RQ 1.1: 
token-level log-probability statistics carry information about reasoning 
quality that is detectable by an external evaluator. Second, the role 
asymmetry first observed in the trajectory analysis is now quantified at 
the correlation level. The Constructor's mean correlation across reasoning 
dimensions ($\bar{\rho} = 0.335$) is roughly twice that of the Auditor 
($\bar{\rho} = 0.177$), and the gap is preserved across every individual 
dimension. Third, the critical-failure AUROC is substantially higher for 
the Constructor ($0.804$) than for the Auditor ($0.634$), suggesting that 
confidence-based diagnostics are most powerful 
for hallucination-style failures and least useful for procedural 
violations whose token-level signature is more subtle.

\begin{table}[t]
\centering
\small
\caption{Preliminary cross-signal results aggregated across rubric-scoring 
experiments. Each cell reports 
the mean of the strongest per-set values, where ordinal-target rows are 
Spearman $\rho$ between the top token-level confidence proxy and the 
LLM-as-judge score for that dimension, and the critical-failure row is 
AUROC. Constructor and Auditor refer to the rubric-domain instantiations of the abstract debate roles defined in 
Section~\ref{sec:meth}.}
\label{tab:rq-summary}
\begin{tabular}{l c c}
\toprule
Reasoning quality dimension & Constructor & Auditor \\
\midrule
Instruction Following ($\rho$) & 0.384 & 0.170 \\
Justification Quality ($\rho$) & 0.319 & 0.231 \\
Evidence Grounding ($\rho$)    & 0.289 & 0.103 \\
Aggregate Score ($\rho$)       & 0.350 & 0.202 \\
\midrule
Critical Failure (AUROC)       & 0.804 & 0.634 \\
\midrule
Mean role correlation $\bar{\rho}$ & \textbf{0.335} & 0.177 \\
Asymmetry ratio (C/A)              & \multicolumn{2}{c}{$1.89\times$} \\
\bottomrule
\end{tabular}
\end{table}

Taken together, the trajectory and correlation analyses point in the same 
direction: confidence signals reflect reasoning quality more reliably for 
supportive than for adversarial argumentation, the most informative 
regions of the response lie at its boundaries rather than its middle, and 
critical failures leave detectable token-level signatures whose strength 
depends on the failure mode. At the same time, these results reflect a 
single domain (rubric-based scoring), a single model family 
(GPT-4o-mini for generation, GPT-5-mini for judging), and a single judge 
instance, and therefore cannot speak to whether confidence--quality 
alignment generalizes across reasoning types, whether the 
observed correlations transfer to settings where the ground truth is 
verifiable rather than rubric-defined, or whether the 
LLM judge itself is consistent across model families and protocols 
.


\section{Conclusion}

This paper proposes a framework for evaluating intermediate reasoning in
multi-agent debate by jointly analyzing token-level log-probabilities,
LLM-as-judge rubric scores, and task accuracy across rubric scoring,
mathematical reasoning, and factual question answering. Experiments in
the rubric domain reveal a consistent four-phase confidence trajectory
and a substantial role asymmetry: confidence aligns with judged quality
roughly twice as strongly for supportive (Constructor) as for adversarial
(Auditor) reasoning, with a parallel gap in critical-failure detection.
Extending this analysis to mathematical reasoning and factual question
answering, alongside multi-judge consistency checks and divergence-based
diagnostics, will inform the design of more interpretable and
trustworthy debate architectures.

\section*{Acknowledgments}
This paper is based upon work supported by the National Science Foundation under Grant No. 2315294.

\section*{Limitations}

Several limitations constrain the interpretation and generalizability 
of the findings presented in this work. Although the proposed framework 
is designed to operate across multiple domains, the current empirical 
results remain narrow in scope, covering only a small set of benchmark 
tasks, datasets, and debate configurations. While rubric-based scoring, 
mathematical reasoning, and factual question answering capture distinct 
forms of reasoning behavior, they do not represent the full range of 
environments in which multi-agent systems are deployed, such as 
long-context reasoning, multimodal tasks, retrieval-heavy workflows, 
code generation, planning, or interactive decision-making. All reported 
experiments also use a single model family for both generation  and judging, leaving open whether the 
observed dynamics reflect properties of debate itself or of a specific 
decoder.

The debate architecture is also intentionally simplified, relying on 
two agents and a single Synthesizer with fixed interaction order and 
limited debate depth. Real-world multi-agent systems often involve 
iterative refinement, retrieval augmentation, memory mechanisms, and 
more complex communication structures that may produce substantially 
different confidence dynamics.

Finally, intermediate reasoning traces and token-level confidence 
signals may not faithfully reflect the internal computation responsible 
for a model's final answer, meaning that both log-probability 
trajectories and LLM-as-judge evaluations could capture post hoc 
rationalizations rather than genuine reasoning processes. The observed 
relationships between confidence, judged reasoning quality, and 
downstream accuracy should therefore be interpreted cautiously, and 
future work should expand the framework across broader datasets, 
models, prompting paradigms, and interaction structures.

\bibliography{custom}

\appendix
\section{Agent Prompt Templates}
\label{app:prompts}

This appendix provides the system instructions for all agents in the 
multi-agent debate framework. All prompts are implemented as templates 
rendered at runtime using a shared context dictionary containing the 
task input $x$, task context $c$, and valid output range where 
applicable. The Constructor and Auditor receive only their own turn; the 
Synthesizer receives the full debate transcript. The LLM-as-judge 
meta-evaluator receives the original agent system prompt, the task 
context, and the agent response being evaluated.

Prompts were developed through iterative piloting to enforce role 
adherence, prevent premature answer generation, and maintain output 
consistency across domains. The Constructor and Auditor share a common 
template structure---role preamble, behavioral constraints, task context, 
prior debate turn (Auditor only), and output format---with only the role 
preamble varying across agents and domains.

\subsection{Constructor Agent}

The Constructor produces the primary argument or solution for a given 
task instance, committing to a direction and developing it with explicit 
reasoning and evidence. It is prohibited from producing a final task 
output in the rubric scoring and QA domains; in the math domain it must 
show a complete derivation. The domain-specific persona is injected via 
\texttt{\$ROLE\_PREAMBLE}.

\begin{figure*}[h]
\centering
\begin{tcolorbox}[title=\textbf{Constructor Agent System Prompt}, 
  colback=blue!3, colframe=blue!60]
\textbf{Role Preamble}\\
\$ROLE\_PREAMBLE

\medskip
\textbf{[Rubric Scoring]:} You are an Advocate. Analyze the essay and 
present evidence-based arguments for how it satisfies the rubric 
expectations for the trait ``\$TRAIT\_NAME''. Focus exclusively on 
supporting evidence. Do not assign a score or discuss weaknesses. 
Reference specific passages from the essay.

\medskip
\textbf{[Mathematical Reasoning]:} You are a Solver. Produce a complete, 
step-by-step solution to the problem. Show all intermediate derivations 
explicitly. State your final answer clearly at the end. Do not skip steps 
or assert results without justification.

\medskip
\textbf{[Factual QA]:} You are a Proponent. Argue for the most 
well-supported candidate answer given the question and available context. 
Cite specific evidence from the provided passages. Explain why competing 
answers are less supported. Do not state a final answer directly.

\medskip
\textbf{Shared Constraints}\\
Organize your response using the following sections: \textit{Claim}, 
\textit{Evidence}, \textit{Reasoning}, \textit{Conclusion}.\\
Do not mix multiple traits, sub-problems, or questions in a single 
response.\\
\$ANON\_CONTEXT
\end{tcolorbox}
\caption{System instructions for the Constructor agent. The 
\texttt{\$ROLE\_PREAMBLE} slot is replaced with the domain-specific 
block at runtime.}
\label{fig:prompt_constructor}
\end{figure*}

\subsection{Auditor Agent}

The Auditor reads the Constructor's output and produces an independent 
critical response. It must not simply restate or paraphrase the 
Constructor's reasoning; in the math domain it must rederive relevant 
steps before issuing a judgment.

\begin{figure*}[h]
\centering
\begin{tcolorbox}[title=\textbf{Auditor Agent System Prompt}, 
  colback=red!3, colframe=red!60]
\textbf{Role Preamble}\\
\$ROLE\_PREAMBLE

\medskip
\textbf{[Rubric Scoring]:} You are a Critic. Identify weaknesses or 
rubric misalignments in the essay for the trait ``\$TRAIT\_NAME''. 
Challenge the Advocate's claims using specific textual evidence distinct 
from that already cited. Do not assign a score or discuss strengths.

\medskip
\textbf{[Mathematical Reasoning]:} You are a Verifier. Independently 
check each step of the Solver's derivation for correctness. If an error 
exists, identify the first incorrect step and provide an alternative 
derivation from that point. Do not confirm or deny the final answer 
without showing your own working.

\medskip
\textbf{[Factual QA]:} You are a Challenger. Question the Proponent's 
evidence and argue for the most strongly supported alternative answer. 
Cite specific passages that the Proponent overlooked or misinterpreted. 
Do not reuse evidence already cited by the Proponent.

\medskip
\textbf{Shared Constraints}\\
Organize your response using the following sections: \textit{Challenge}, 
\textit{Counter-Evidence}, \textit{Reasoning}, \textit{Conclusion}.\\
Do not introduce information outside the provided task context.\\
\$ANON\_CONTEXT
\end{tcolorbox}
\caption{System instructions for the Auditor agent.}
\label{fig:prompt_auditor}
\end{figure*}

\subsection{Synthesizer Agent}

The Synthesizer reads the completed debate transcript and produces the 
final task output. It is held constant across all domains and its output 
is evaluated directly against $y^*$.

\begin{figure*}[h]
\centering
\begin{tcolorbox}[title=\textbf{Synthesizer Agent System Prompt}, 
  colback=gray!5, colframe=black!60]
You are the Synthesizer in a multi-agent debate system. You will receive 
a debate transcript between a Constructor and an Auditor addressing the 
following task.

\medskip
\textbf{[Rubric Scoring]:} Read both arguments and assign a final integer 
score between \$MIN\_POINTS and \$MAX\_POINTS for the trait 
``\$TRAIT\_NAME''. Base your decision on the strength and specificity of 
the evidence presented by both agents.

\medskip
\textbf{[Mathematical Reasoning]:} Read the Solver's solution and the 
Verifier's critique. Determine the correct final answer. If the Verifier 
identified an error, incorporate the corrected derivation. State the 
final answer explicitly.

\medskip
\textbf{[Factual QA]:} Read the Proponent's argument and the 
Challenger's response. Select the best-supported answer from the 
candidates discussed. Justify your selection in one sentence.

\medskip
\textbf{Shared Constraints}\\
Do not introduce new arguments or evidence not present in the transcript.\\
Your output must be a single final answer in the format specified above.\\
\$ANON\_CONTEXT
\end{tcolorbox}
\caption{System instructions for the Synthesizer agent.}
\label{fig:prompt_synthesizer}
\end{figure*}

\subsection{LLM-as-Judge Meta-Evaluator}
\label{sec:meta-evaluator-agent}

The meta-evaluator scores the Constructor's and Auditor's responses 
along three dimensions: instruction following, justification quality, 
and evidence grounding, each on a three-point ordinal scale 
(1~=~Weak, 2~=~Adequate, 3~=~Strong). It also raises a critical issue 
flag for severe failures. Crucially, the meta-evaluator is instructed 
to evaluate only the \emph{quality of the agent's reasoning}, not the 
correctness of the task output or the content of the essay, problem, 
or passage. The evaluation dimensions are intentionally kept consistent 
across domains so that judge scores are comparable across rubric 
scoring, math, and QA.

\begin{figure*}[h]
\centering
\begin{tcolorbox}[title=\textbf{Meta-Evaluator System Prompt}, 
  colback=blue!8, colframe=blue!40!black]
You are a meta-evaluator assessing the reasoning quality of an AI agent 
in a multi-agent debate system.

\medskip
\textbf{Important:} Do NOT evaluate the task answer or judge whether 
the agent's position is correct. Evaluate ONLY the agent's reasoning 
quality, role adherence, and use of evidence.

\medskip
You will receive: (1) the agent's system prompt, (2) the task context 
provided to the agent, and (3) the agent's response. Evaluate across 
three dimensions using the full range of the scale. Avoid defaulting 
to the middle score. Evaluate each dimension independently.

\medskip
\textbf{Dimension 1 --- Instruction Following.}\\
3: Fully maintains role; completes all required output sections; no 
deviations.\\
2: Generally follows instructions with a minor omission or slight 
deviation.\\
1: Major or multiple deviations; neglects important instructions.

\medskip
\textbf{Dimension 2 --- Justification Quality.}\\
3: Multiple claims with clear reasoning; 
claim~$\rightarrow$~evidence~$\rightarrow$~implication structure 
throughout.\\
2: At least one supported claim; reasoning understandable but shallow 
or repetitive.\\
1: Minimal or vague reasoning; assertions without explanation.

\medskip
\textbf{Dimension 3 --- Evidence Grounding.}\\
3: Two or more precise, specific references to the task input (quotes, 
equations, passage spans).\\
2: One clear identifiable reference; remaining claims rely on general 
statements.\\
1: Evidence vague, indirect, or absent.

\medskip
\textbf{Critical Issues Flag.} Set \texttt{critical\_flag = 1} if any 
of the following occur: hallucinated evidence not present in the task 
input, severe internal contradiction, explicit violation of role 
constraints, or incoherent output. Otherwise \texttt{critical\_flag = 0}.

\medskip
\textbf{Output:} Return only a JSON object with fields 
\texttt{instruction\_following}, \texttt{justification\_quality}, 
\texttt{evidence\_grounding}, \texttt{critical\_flag}, 
\texttt{critical\_issues\_description}, and \texttt{reasoning} 
(2--3 sentences explaining the scores).
\end{tcolorbox}
\caption{System instructions for the meta-evaluator agent.}
\label{fig:meta_system_prompt}
\end{figure*}

\begin{figure*}[h]
\centering
\begin{tcolorbox}[title=\textbf{Meta-Evaluator Task Prompt}, 
  colback=green!7, colframe=green!40!black]
\texttt{\# Agent Being Evaluated: \{AGENT\_TYPE\}}\\
\texttt{\# Domain: \{DOMAIN\}}

\medskip
\texttt{\# Agent's System Prompt}\\
\texttt{\{AGENT\_SYSTEM\_PROMPT\}}

\medskip
\texttt{\# Task Context Provided to the Agent}\\
\texttt{\{AGENT\_USER\_PROMPT\}}

\medskip
\texttt{\# Agent's Response}\\
\texttt{\{AGENT\_RESPONSE\}}

\medskip
Evaluate this agent's response using the 3-point scale for each 
dimension and set \texttt{critical\_flag} to 0 or 1. Output ONLY the 
JSON object.
\end{tcolorbox}
\caption{Task prompt provided to the meta-evaluator at runtime.}
\label{fig:meta_task_prompt}
\end{figure*}

\end{document}